# AD-Aligning: Emulating Human-like Generalization for Cognitive Domain Adaptation in Deep Learning

1st Zhuoying Li*
Department of Computer Science
Johns Hopkins University
Baltimore, US
*Corresponding author: zli181@jhu.edu

2nd Bohua Wan
Department of Computer Science
Johns Hopkins University
Baltimore, US
bwan2@jhu.edu

3rd Cong Mu
Department of Computer Science
Johns Hopkins University
Baltimore, US
cmu2@jhu.edu

4th Ruzhang Zhao
Department of Computer Science
Johns Hopkins University
Baltimore, US
rzhao@jhu.edu

5th Shushan Qiu
Department of Electrical and Computer Engineering
University of Houston
Houston, US
sqiu3@cougarnet.uh.edu

6th Chao Yan
Department of Electrical and Computer Engineering
Northeastern University
Sunnyvale, US
yan.chao@northeastern.edu

*Abstract*—Domain adaptation is pivotal for enabling deep learning models to generalize across diverse domains, a task complicated by variations in presentation and cognitive nuances. In this paper, we introduce AD-Aligning, a novel approach that combines adversarial training with source-target domain alignment to enhance generalization capabilities. By pretraining with Coral loss and standard loss, AD-Aligning aligns target domain statistics with those of the pretrained encoder, preserving robustness while accommodating domain shifts. Through extensive experiments on diverse datasets and domain shift scenarios, including noise-induced shifts and cognitive domain adaptation tasks, we demonstrate AD-Aligning's superior performance compared to existing methods such as Deep Coral and ADDA. Our findings highlight AD-Aligning's ability to emulate the nuanced cognitive processes inherent in human perception, making it a promising solution for real-world applications requiring adaptable and robust domain adaptation strategies.

## I. INTRODUCTION

### A. Background

Generalization across domains is a pivotal aspect of AI cognitive research, particularly in the context of domain adaptation [1]. Understanding how models can effectively adapt to different domains has emerged as a highly active area of investigation [2]. Researchers have employed various strategies to tackle the challenge of domain shift, where the distribution of data differs between training and testing phases [3]. However, much of the existing work primarily focuses on addressing domain shifts within the objective truth—instances where the task remains consistent across domains, albeit with varying environmental conditions [4].

Consider a scenario where a model is trained to identify bears in photos taken during the day but is tested on images captured at night. This exemplifies a common domain shift problem, where the model must generalize its understanding to accommodate changes in lighting conditions. While such challenges are significant, they represent only one facet of domain adaptation [5].

A crucial yet often overlooked aspect is cognitive domain shift, where humans possess the ability to generalize across disparate concepts despite variations in presentation [6]. For instance, if individuals are trained to recognize polar bears but are later tested on images of brown bears, they can seamlessly extend their understanding to identify both as types of bears. This cognitive flexibility presents a formidable challenge for artificial intelligence (AI) systems, as they struggle to emulate the nuanced cognitive processes inherent in human perception [7].

In this paper, we address this gap in research by investigating cognitive domain shift and its implications for machine learning models. We argue that understanding and replicating human-like generalization abilities are essential for developing robust AI systems capable of adapting to diverse and evolving environments. Through empirical analysis and theoretical in-sights, we aim to shed light on the complexities of cognitive domain shift and pave the way for more comprehensive solutions in domain adaptation research [8].

## B. Related Work

Machine learning (ML) has been a focal point of research and innovation in recent years, with numerous influential papers contributing to its advancement. For instance, Liu et al. introduced a method for influence pathway discovery on social media [9]. Additionally, Li et al. proposed a Contextual Hourglass Network for semantic segmentation of high-resolution aerial imagery [10] Furthermore, Li et al. presented a technique for deception detection using bimodal convolutional neural networks, highlighting the importance of leveraging multiple modalities for domain adaptation tasks [11].

Computer vision, a vital field in modern technology, enables machines to comprehend visual data akin to human perception. Recent advancements, such as Wang et al.'s research on image recognition using multimodal deep learning [12], Shree et al.'s patent about Image analysis [13] and innovative techniques for enhanced image capture by Castillo et al. [14] have significantly impacted the field. These developments underscore the growing importance of computer vision across diverse domains, from healthcare to industrial automation.

Domain adaptation, the process of transferring knowledge from a source domain to a target domain with differing distributions, has garnered significant attention in the computer vision community. Numerous approaches have been proposed to address this challenge, ranging from traditional methods to more recent deep learning-based techniques. Traditional domain adaptation methods often rely on minimizing the discrepancy between source and target domains using techniques such as Maximum Mean Discrepancy (MMD) and Kernel Mean Matching (KMM) [15]. Deep learning has revolutionized domain adaptation by leveraging the representational power of deep neural networks to learn domain-invariant features. Notable deep learning-based methods include Deep Coral [4], which aligns second-order statistics to bridge domain gaps, Adversarial Discriminative Domain Adaptation (ADDA) [1], which employs adversarial training to learn domain-invariant representations, and Self-supervised agent learning [16], which learns domain-invariant representations through a self-supervised learning framework.

These methods have demonstrated promising results in various domains and have paved the way for further advancements in domain adaptation research. In recent years, various approaches have been proposed to address similar challenge. Recent works have explored novel techniques such as meta-learning [17]. to enhance the adaptability and robustness of domain adaptation models. Additionally, attention has been given to the challenges of unsupervised, semi-supervised, and multi-source domain adaptation scenarios [18] [19].

## II. METHODOLOGY

### A. Benchmark Models

*1) Deep Coral:* The Deep Coral model [4], introduced by Sun et al. in 2016, enhances domain adaptation by aligning source and target distributions in a deep feature space via correlation alignment. This approach improves adaptation performance across diverse tasks, making Deep Coral a fundamental method in creating robust and generalizable deep learning systems.

*2) ADDA:* The Adversarial Discriminative Domain Adaptation (ADDA) [1] model excels in unsupervised domain adaptation by minimizing domain discrepancy and maximizing task-specific discrimination through adversarial training. It learns domain-invariant representations, enabling effective adaptation without labeled target domain data. ADDA's innovative framework and effectiveness make it a widely used method for addressing domain shift in various applications.

### B. AD-Aligning

We propose an unsupervised model that integrates the advantages of the Adversarial Discriminative model with the alignment of source and target domain distributions. In the adaptation phase, we receive source images $X_s$ with corresponding labels $Y_s$ exhibiting a source domain distribution $p_s(x,y)$, alongside target images $X_t$ characterized by a target distribution $p_t(x,y)$, and unseen target images $X_{ut}$ with an unknown distribution. The primary objective of domain adaptation classification is to obtain a target representation $M_t$ and classification $C_t$ capable of categorizing both $X_t$ and $X_{ut}$ into K classes.

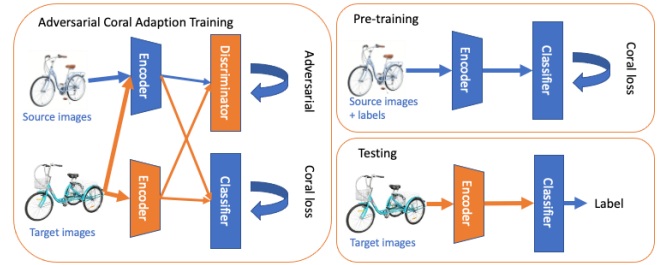

Figure 1. An illustration of our proposed method combining Deep Coral and ADDA. Blue and orange arrows denote data flows of source and target domain respectively.

The model structure is illustrated in Figure 1. During the pre-training phase, source images are fed into an encoder to obtain their representation $M_s$. The classifier $C$ is then trained using the source representation and labels. Following pre-training, we acquire an encoder capable of encoding the source representation and a classifier $C$. Subsequently, in the adversarial adaptation phase, only source and target images are provided without labels. The objective in this phase is to obtain a target encoder that generates a target representation $M_t$ similar to the source representation $M_s$. To achieve this, we freeze the source image encoder and target classifier. The initial status of the target encoder is copied from the source encoder, and a discriminator is employed to work adversarially with the target encoder to train it in an unsupervised manner. Finally, the target encoder and source classification model are used to classify the target images.

In both the pre-training and adversarial training phases, we employ standard supervised loss augmented with coral loss:

$$l_{cls} = l_{class} + \sum_{i=1}^{t} \lambda_i l_{coral} \qquad (1)$$

where t represents the number of coral loss layers in the network, and λ is a weight parameter that balances the contribution of the coral loss. The classification losses ($l_{class}$) and coral losses ($l_{coral}$) are expressed as follows:

$$l_{class}(X_s, Y_t) = \mathbb{E}_{(X_s, Y_t)} - \sum_{k=1}^{t} \log C(M_s(x_s)) \qquad (2)$$

$$l_{coral} = \frac{1}{4d^2} ||Co_s - Co_t||_F^2 \qquad (3)$$

where $d$ denotes the dimensionality of the deep layer, *Cos (Cot)* represents the feature covariance matrices, and $||.||_2^F$ denotes the squared matrix Frobenius norm.

In the adversarial adaptation task, a domain discriminator *D* is introduced to discern whether the representation originates from the source or target domain. The adversarial component also incorporates standard supervised loss, formulated as:

$$l_{advD} = -\mathbb{E}_{X_s}[\log D(M_s(x_s))] - \mathbb{E}_{X_t}[\log(1 - D(M_s(x_s)))] \qquad (4)$$

Given that the target domain lacks labels, minimizing the disparity between source and target representations poses a challenge. During training of the adversarial network, the encoder is trained using the standard loss function with inverted labels [10], yielding the encoder loss:

$$l_{advE} = -\mathbb{E}_{X_t}[D(M_s(x_s))] \qquad (5)$$

AD-Aligning entails unconstrained optimization aimed at minimizing the combined losses $l_{cls}$, $l_{advD}$, and $l_{advE}$.

Adversarial loss ensures the target encoder generates representations indistinguishable from the source representations, thereby aligning the source and target domain distributions.

### C. Data Processing

To conduct our experiments, it is imperative to procure datasets comprising both source and target images exhibiting domain shift. We meticulously curate three pairs of datasets to ensure the requisite conditions for our experiments are met.

1) Noise Dataset: Noise represents a prevalent manifestation of domain shift, imparting notable alterations to the texture and characteristics of images. Notably, images captured under low-light conditions, such as during nighttime, often exhibit heightened levels of noise compared to those captured in well-lit environments. In our experimental setup, we delineate the original, noise-free image as the source, while the image subjected to noise augmentation serves as the target. This approach effectively simulates domain shifts corresponding to various times of the day. Specifically, our noise dataset encompasses five distinct types of perturbations introduced into the Tiny-16-Class-ImageNet dataset: uniform noise, salt-and-pepper noise, rotation, high-pass, and low-pass filters. A visual depiction of domain shift induced by noise is provided in Figure 2.

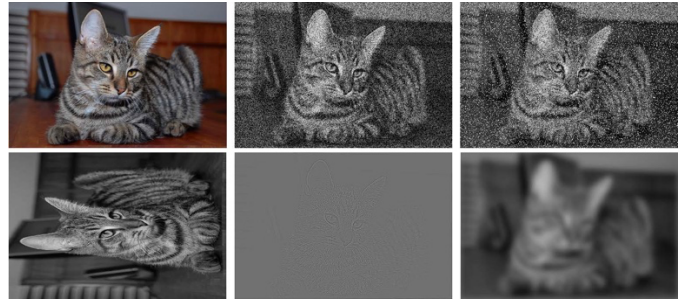

Figure 2. Example images across different domains. Top row from left to right: source, uniform noise, salt-and-pepper noise. Bottom row from left to right: rotation, high-pass, low-pass. Imagemanipulations follow the procedure in [20].

2) MNIST-USPS: The MNIST-USPS dataset amalgamates handwritten digit images sourced from two disparate origins: MNIST and USPS. MNIST comprises standardized 28x28 grayscale images depicting digits 0 through 9, while USPS encompasses handwritten digits gleaned from postal mail envelopes. This dataset assumes a pivotal role as an evaluative benchmark for domain adaptation algorithms, owing to the substantial disparities in style, dimensions, and variability between the source and target domains. Such diversity renders it an exemplary testbed for gauging the efficacy of domain adaptation methodologies in real-world contexts.

3) Generalized ImageNet: The Tiny-16-Class-ImageNet encompasses 16 general classes, with multiple subclasses within each class. Through manual integration of all datasets, we partitioned the dataset into source, target, and unseen-target subsets based on subclasses. For instance, all images depicting brown bears were categorized into the source subset, while half of the images featuring black bears were allocated to the target subset, and the remaining black bear images were assigned to the unseen-target subset. The ground truth for these images uniformly identifies them as belonging to the bear category. This categorization process was informed by human cognition. Thus, we curated the Generalized ImageNet dataset, intending to assess the model's efficacy in Cognitive Domain Adaptation.

### III. EXPERIMENTS

#### A. Deep Coral

As a benchmark, we configure Deep Coral [4] to align the second-order statistics within the final layer of the backbone network by incorporating a coral loss. Recognized for its efficacy and remarkable versatility, this method provides a robust foundation for comparison. Significantly, we enhance the framework by substituting Deep Coral's backbone with a ResNet-50 model pretrained on the ImageNet dataset, thereby leveraging established representations to enhance performance. We adopt the same SGD hyperparameters as outlined in [4]. The λ parameter, controlling the weight of the coral loss, remains consistent with [4], except on the MNIST-USPS dataset, where we set $\lambda = 1 - \frac{epoch}{num_epoch}$.

## B. ADDA

To establish a benchmark, we adhere to the methodology of Adversarial Discriminative Domain Adaptation (ADDA) by initially acquiring a discriminative representation using source domain data. Subsequently, we utilize a domain-adversarial loss to train an additional encoder, which maps the target domain to the source domain. For our implementation, we employ ResNet-50 as the backbone for the encoder, complemented by a three-layer Multi-Layer Perceptron (MLP) acting as the discriminator, with a hidden size of 1024. The Adam optimizer is utilized, with parameters $b1 = 0.5$ and $b2 = 0.999$. We set the learning rate to $0.0002$ and utilize a batch size of $32$. During the adaptation stage, updates to the target encoder are performed every four steps.

## C. AD-Aligning

We employ Coral loss in conjunction with standard loss during the pretraining phase of AD-Aligning to align the second-order statistics of target domains between the classification outputs of the fixed pretrained encoder and the AD-Aligning trained target encoder. The overall architecture is depicted in Figure 1. The blue encoder and classifier are pretrained and remain fixed throughout the experiments. Our experimentation reveals that the vanilla Adversarial Discriminative component compromises the pretrained encoder due to the inadequately trained discriminator. To optimize the utilization of the pretrained encoder initialization while ensuring that the target encoder generates similar features for the target and source domains, we utilize the coral loss solely to align the classification output of the AD-Aligning trained encoder with that of the fixed pretrained encoder, gradually reducing the weight of the coral loss over time.

## IV. DISCUSSION

### A. Compare ADDA with Deep CORAL on Noise Dataset

We conducted experiments on the Noise Dataset to evaluate the performance of ADDA and Deep CORAL in handling domain shifts induced by five types of noise. Models were trained using the source dataset and one noise target dataset. Their performance was assessed across all five noise domains.

The experimental results, illustrated in Figure 3, showcase the improvements made by ADDA and Deep CORAL on the target domain. The classification accuracies for different domains are presented as percentages. Model M0 was solely trained on the source domain, while models M1 to M5 were adapted to one target domain using ADDA. Similarly, models M6 to M10 were adapted to one target domain using Deep CORAL. The red rectangle highlights the target domain on which each model was trained. Notably, the best results for each domain and method are highlighted in bold blue.

In our analysis, Deep CORAL consistently outperforms ADDA across various target domains, with the exception of the High-Pass domain. The discrepancy in performance on the High-Pass domain can be attributed to the significant domain shift present between High-Pass and the other domains. This challenge underscores the limitations of Deep CORAL in handling extreme domain shifts, where features learned in the source domain may not generalize well to the target domain.

Despite this, Deep CORAL demonstrates superior generalizability to previously unseen domains compared to ADDA. This enhanced generalizability can be attributed to the minimal alteration of the encoder by Deep CORAL, particularly as the encoder is pretrained on the ImageNet dataset without the inclusion of added noises. This pretrained initialization likely contributes to the model's ability to capture robust and transferable features, thereby enhancing its performance across diverse domains.

| | $M_0$ | $M_1$ | $M_2$ | $M_3$ | $M_4$ | $M_5$ | $M_6$ | $M_7$ | $M_8$ | $M_9$ | $M_{10}$ |
|---|---|---|---|---|---|---|---|---|---|---|---|
| Rotation (90º) | 50.89 | **53.78** | 23.14 | 11.91 | 1.53 | 3.40 | **85.42** | 62.71 | 68.89 | 81.26 | 81.56 |
| Uniform Noise (0.35) | 35.06 | 25.78 | **68.93** | 58.93 | 1.53 | 2.97 | 80.75 | **88.20** | 83.26 | 79.74 | 76.97 |
| Salt-and-Pepper Noise (0.2) | 25.02 | 24.34 | 36.46 | **65.27** | 1.53 | 1.57 | 71.03 | **79.66** | 76.20 | 64.66 | 69.98 |
| High-Pass (std=0.7) | 23.64 | 22.11 | 5.76 | 1.82 | **57.41** | 22.62 | 18.72 | 21.97 | 17.30 | 21.13 | **24.87** |
| Low-Pass (std=7) | 37.10 | 50.63 | 6.55 | 3.19 | 1.53 | **51.27** | 48.71 | 19.64 | 24.66 | 59.67 | **71.78** |

Figure 3. Classification accuracy in percent for different domains. Model M0 is only trained on source domain. Models M1 to M5 are adapted on one target domain (in red rectangle) via ADDA. M6 to M10 are similar except with Deep Coral. Best results for each domain and method are bold in blue.

### B. Performance on Cognitive Domain Adaptation

We conducted tests on more challenging datasets to evaluate the models' performance in cognitive domain adaptation, aiming to assess their ability to generalize across disparate concepts despite variations in presentation. The datasets used for this evaluation were MNIST-USPS and Generalized ImageNet. To increase the difficulty of the task, uniform noise (0.5) was added to the Generalized ImageNet target domain.

MNIST-USPS was chosen to assess the models' basic ability in handling domain shifts related to objective truth, while Generalized ImageNet was used to evaluate their performance in cognitive domain adaptation.

TABLE I. COMPARE MODELS ON DOMAIN SHIFT DATASET

| Dataset | Target Domain Accuracy % | | | |
|---|---|---|---|---|
| | ResNet-50 | ADDA | Deep Coral | AD-Aligning |
| MNIST-USPS | 25.13 | 89.40 | 54.30 | **94.56** |
| Generalized ImageNet[a] | 23.15 | 48.32 | 73.52 | **77.69** |

a. target domain set with uniform noise (0.5)

Table I. presents the results of different models on MNIST-USPS and Generalized ImageNet. ResNet-50 exhibited limited capability in handling domain shifts, performing slightly better than random guessing but showing poor accuracy in the target domain. ADDA demonstrated good performance on MNIST-USPS but performed poorly on Generalized ImageNet, indicating its effectiveness in objective truth domain shift but limitations in cognitive domain adaptation.

Conversely, Deep Coral showed better performance in cognitive domain adaptation, possibly due to its utilization of correlation alignment, which effectively emulates the nuanced cognitive processes inherent in human perception. However, the reliability of Deep Coral's results is questionable, given its poor performance on ADDA.

Our proposed model, AD-Aligning, outperformed other settings in both MNIST-USPS and Generalized ImageNet tests. The notable improvements in performance validate the effectiveness of our novel modifications and designs, highlighting the potential of AD-Aligning as a robust solution for domain adaptation tasks.

*C. Performance on Unseen Domain Adaptation*

Not all target domains are readily available during the training phase, necessitating an investigation into the model's performance when faced with unseen target domains. To validate this, we utilize the Generalized ImageNet dataset. To intensify the challenge, uniform noise (0.5) is added to the target domain. We deliberately abstain from training ResNest50-ImageNet, a pretrained model, to serve as a control model.

TABLE II. COMPARE MODELS ON UNSEEN DATASET

| Model | Source | Target | Accuracy% |
|---|---|---|---|
| ResNet-50 | train | None | 5.34 |
| ResNet50-ImageNet | None | None | 13.14 |
| Deep Coral | train | val[a] | 38.37 |
| AD-Aligning | train | val[a] | **52.96** |

a. Validation set with uniform noise (0.5)  b. Using Generalized ImageNet Dataset

The results are presented in Table II. ResNet-50 demonstrates an inability to adapt to the target domain, displaying signs of overfitting on the source domain and performing even worse than the untrained ResNest50-ImageNet. Conversely, Deep Coral exhibits notably superior performance compared to ResNet-50. AD-Aligning's performance on the unseen and untrained target domain significantly outperforms other methods, as evidenced in Table 2. Given the inherent difficulty of adapting to unseen target domains, the achieved accuracy underscores the robustness of the model.

V. CONCLUSION

Our experimental investigation into domain adaptation methods, including Deep Coral, ADDA, and our proposed AD-Aligning approach, has provided valuable insights into their effectiveness across various domains and scenarios. Deep Coral demonstrates superior generalizability to unseen domains compared to ADDA, although it struggles with extreme domain shifts such as in the High-Pass domain. On the other hand, ADDA performs well on objective truth domain shift tasks but exhibits limitations in cognitive domain adaptation. Our novel AD-Aligning method showcases significant improvements over existing approaches, particularly in addressing the challenges of unseen domain adaptation. Its robust performance across diverse datasets underscores its potential as a versatile and effective solution for domain adaptation tasks. Overall, our findings highlight the importance of considering the nuances of domain shifts and the need for adaptable and robust adaptation methods to address real-world challenges effectively.